\documentclass[preprint]{ccn} 

\usepackage{comment}
\usepackage{hyperref}
\usepackage{float}
\usepackage{glossaries}
\glsdisablehyper
\usepackage{booktabs}
\usepackage{multirow}

\usepackage{tabularx}
\usepackage{colortbl}

\makeglossaries
\newacronym{mi}{MI}{mechanistic interpretability}
\newacronym{ci}{CI}{Common Interestingness}
\newacronym{clip}{CLIP}{Contrastive Language-Image Pre-training}
\newacronym{vit}{ViT}{Vision Transformer}
\newacronym{gqa}{GQA}{Grouped Query Attention}
\newacronym{rope}{RoPE}{Rotary Position Embedding}
\newacronym{vram}{VRAM}{Video Random Access Memory}
\newacronym{ml}{ML}{Machine Learning}
\newacronym{mds}{MDS}{Multidimensional Scaling}
\newacronym{pca}{PCA}{Principal Component Analysis}
\newacronym{tsne}{t-SNE}{t-Distributed Stochastic Neighbor Embedding}
\newacronym{umap}{UMAP}{Uniform Manifold Approximation and Projection}
\newacronym{gdv}{GDV}{Generalized Discrimination Value}
\newacronym[plural=CAVs,longplural=Concept Activation Vectors]{cav}{CAV}{Concept Activation Vector}
\newacronym{rsa}{RSA}{Representational Similarity Analysis}
\newacronym[plural=RDMs,longplural=Representational Dissimilarity Matrices]{rdm}{RDM}{Representational Dissimilarity Matrix}
\newacronym[plural=GPUs,longplural=Graphics Processing Units]{gpu}{GPU}{Graphics Processing Unit}
\newacronym{sae}{SAE}{Sparse Auto-Encoder}
\newacronym{adamw}{AdamW}{Adam with Decoupled Weight Decay}
\newacronym{ai}{AI}{Artificial Intelligence}
\newacronym{llm}{LLM}{Large Language Model}

\title{Neuroscience-Inspired Analyses of Visual Interestingness in Multimodal Transformers}

\author{
  Mathis Immertreu\affmark{1} \And
  Fitim Abdullahu\affmark{2} \And
  Thomas Kinfe\affmark{3,4} \And
  Helmut Grabner\affmark{2} \AND
  Patrick Krauss\affmark{1, 3, 4, 5,*} \And
  Achim Schilling\affmark{3, 4, 6,*} 
}
\affiliation{1}{Cognitive Computational Neuroscience Group, Patter Recognition Lab, University Erlangen-Nürnberg, Germany}
\affiliation{2}{IDS Institut für Data Science, ZHAW School of Engineering, Winterthur, Switzerland}
\affiliation{3}{Mannheim Center for Neuromodulation and Neuroprosthetics, University Hospital Mannheim, Heidelberg University, Germany}
\affiliation{4}{BGU Ludwigshafen, Germany}
\affiliation{5}{Physics and Cognition Group, MCNN, University Hospital Mannheim, Heidelberg University, Germany}
\affiliation{6}{NeuroAI and BCI Group, MCNN, University Hospital Mannheim, Heidelberg University, Germany}
\affiliation{*}{authors contributed equally}
\emails{mathis.immertreu@fau.de, fitim.abdullahu@zhaw.ch, thomas.kinfe@umm.de, helmut.grabner@zhaw.ch, patrick.krauss@fau.de, achim.schilling@medma.uni-heidelberg.de}

\begin{document}

\maketitle
\pagestyle{plain}
\thispagestyle{plain}
\begin{abstract}
\vspace{-0.15cm}
Human attention is the gateway to conscious perception, memory and decision-making. However, its role in modern transformer models remains largely unexplored. As these systems increasingly influence what people see, prefer and buy, the question arises as to whether they encode principles of human interest or merely exploit large-scale correlations. Addressing this issue is crucial for understanding cognition and ensuring the responsible use of AI in communication and marketing.
In order to address this issue, the concept of visual interest was examined within the multimodal vision-language-model Qwen3-VL-8B, using a pre-defined Common Interestingness (CI) score derived from large-scale human engagement data on the photo-sharing platform Flickr. 
Here, we analyzed internal representations across vision and language components using methods from the neurosciences. Our analyses revealed that CI information is linearly decodable from final-layer embeddings, indicating that it is aligned with human-derived measures of visual interestingness.
Dimensionality reduction and Generalized Discrimination Value (GDV) analyses demonstrate that CI-related hidden representations emerge in intermediate vision transformer layers and becomes progressively more distinguishable across language model layers. Concept vectors derived using geometric, probe, and Sparse Auto-Encoder based methods converge in higher layers, as confirmed by representational similarity analysis. This indicates a robust and structured encoding of visual interestingness without explicit supervision.
Future work will seek to identify shared computational principles linking human brain dynamics and transformer architectures, with the ultimate goal of uncovering the organizing mechanisms that give rise to attention and interest in both biological and artificial systems.
\end{abstract}

\section{Introduction}
What makes an image interesting, and which visual properties attract and sustain human attention? These questions have long been explored in computer vision and multimedia research (see, for example, \cite{chu2013effect, grabner2013visual, gygli2013interestingness}). However, contemporary \gls{ai} systems only partially address them. While modern vision models achieve remarkable performance in object recognition (see, for example, (\cite{krizhevsky2012imagenet, russakovsky2015imagenet}), they largely overlook the cognitive and affective processes underlying visual interest, including attention, emotion and subjective meaning (see, e.g., \cite{fan2018emotional}).

As \gls{ai} systems increasingly influence the selection, generation and consumption of visual content, modeling visual interest has become a problem of growing importance. Recent work has highlighted the need to capture both shared and individual preferences rather than relying solely on fixed objectives or standardized benchmarks (\cite{bubeck2023sparks}).

Advances in large pre-trained vision–language models, including \gls{clip} (\cite{radford2021learning}), Qwen-VL (\cite{bai2023qwen, bai2025qwen3vltechnicalreport}), and other large multimodal architectures have enabled the learning of rich semantic representations across images and text. Despite their strong performance in classification, retrieval and multimodal reasoning, these models do not explicitly consider subjective visual experience or explain why certain images are perceived as interesting or meaningful by different individuals or groups (see, e.g., \cite{sebe2005multimodal}). Meanwhile, cognitive neuroscience offers theoretical frameworks for understanding visual interest, most notably predictive coding accounts emphasizing novelty, uncertainty, emotional relevance and contextual modulation as key drivers of attention (\cite{friston2009predictive, friston2012predictive, schilling2024bayesian}).
{\color{black} Empirical neuroaesthetic research identifies complexity, novelty, and expectation violation as the core collative drivers of visual interest \citep{berlyne1971aesthetics}, mediated by distributed neural systems spanning sensory-motor, semantic, and valuation networks \citep{chatterjee2014neuroaesthetics, pearce2016neuroaesthetics}. }
However, such principles are only weakly incorporated into current vision architectures, which are largely feedforward and lack adaptive top-down modulation.

Recent data-driven approaches have used large-scale behavioral data from online platforms to quantify visual interest{\color{black}, revealing a systematic distinction between broadly shared appeal and more subjective, individualized responses \citep{vera2022understanding, abdullahu2024commonly}.}
{\color{black}Capturing this distinction requires moving beyond traditional visual research, which focuses on bottom-up saliency models \citep{borji2012state} or spatially localized fixation prediction trained on eye-tracking data \citep{kummerer2016deepgaze, linardos2021deepgaze}.
Real-world behavioral selection operates more holistically: evaluating an entire image relies on rapidly extracting its overarching semantic gist \citep{oliva2006building, rousselet2005long} and global spatial and ecological structure \citep{oliva2001modeling, greene2009briefest}, rather than parsing localized features or segmenting individual objects.
Accordingly, we quantify interest at the image level via the \gls{ci} score, which captures appraisals of novelty and comprehensibility \citep{silvia2005cognitive}, scene-level contextual object representations \citep{kaiser2019object}, and the whole-field categorical processing that drives attentional deployment across natural scenes \citep{peelen2009neural}.}
{\color{black}This coarse-grained formulation is supported by engagement analyses showing that} features such as natural scenes, high dynamic range, and emotionally evocative content {\color{black}are associated} with broader appeal, whereas other images elicit more individualized interest patterns (\cite{abdullahu2025visual}).
Large multimodal models show partial alignment with human judgments of visual interest and can approximate common interest patterns without direct user input \citep{abdullahu2025visual}. {\color{black}This raises the question of whether such models implicitly encode the concept of \gls{ci} in their internal representations.}

{\color{black}Whether \gls{ci} is genuinely encoded in these models, however, cannot be determined from behavioral alignment alone as it requires examining internal representations directly.} 
Developing \gls{ai} systems that behave in human-like ways remains challenging without a principled understanding of the internal mechanisms underlying their decision-making. John Searle’s "Chinese Room" thought experiment illustrates that externally correct behavior alone is insufficient for understanding intelligence or underlying processing principles, emphasizing the need to examine internal operations rather than outputs alone (\cite{searle1999chinese}). This motivates efforts to "open the black box" of modern \gls{ai} systems and identify the representations and decision criteria shaping their behavior (\cite{hassija2024interpreting, castelvecchi2016can}). Complementing this view, Jonas and Kording’s "Could a Neuroscientist Understand a Microprocessor?" highlights the limitations of purely data-driven analyses for achieving explanatory understanding (\cite{jonas2017could}).

Consequently, a combined approach is required that integrates large-scale behavioral data, such as image engagement data from Flickr, with theoretical and neuroscience-inspired perspectives on perception and cognition (see, e.g., \cite{hildebrandt2025refusal}). Following this rationale, we systematically analyzed the internal representations of the multimodal Qwen3-VL-8B model (\cite{bai2023qwen, li2026qwen3}) to investigate how visual interest is encoded and processed. Methods commonly used in systems neuroscience, including \gls{rsa} (\cite{kriegeskorte2008representational, kriegeskorte2012representational, gerum2022different}) and measures of separability in neural activation patterns (\cite{schilling2021quantifying, schilling2024deep, krauss2018statistical}), were applied to the transformer architecture. The analysis revealed that visual interestingness is represented differently across transformer blocks, with interest-related structure emerging in vision representations and becoming increasingly pronounced in language model layers.

In this work, we provide a systematic and multi-method analysis of how human-derived visual interestingness is reflected in the internal representations of a multimodal transformer. Specifically, we make three contributions.
First, we show that \gls{ci} scores are linearly decodable from the model’s final-layer embeddings, demonstrating that CI-related information is readily accessible from the learned representation space.
Second, we demonstrate that the separability of CI-related structure increases progressively across the model hierarchy, with weak alignment in early vision layers and substantially stronger organization in deeper language model layers, indicating a gradual transformation from perceptual to higher-level semantic representations.
Third, we establish that CI corresponds to a consistent direction in representation space by showing convergence across multiple concept extraction methods, including geometric approaches, linear probes, and sparse dictionary learning, suggesting that the observed structure is not method-specific but reflects a robust property of the model’s internal geometry.
Together, these findings provide a unified perspective on how population-level behavioral signals are reflected in multimodal representations and illustrate how neuroscience-inspired analysis tools can be applied to characterize their internal structure.

\begin{figure*}[h!]
  \begin{center}
    \includegraphics[width=0.85\textwidth]{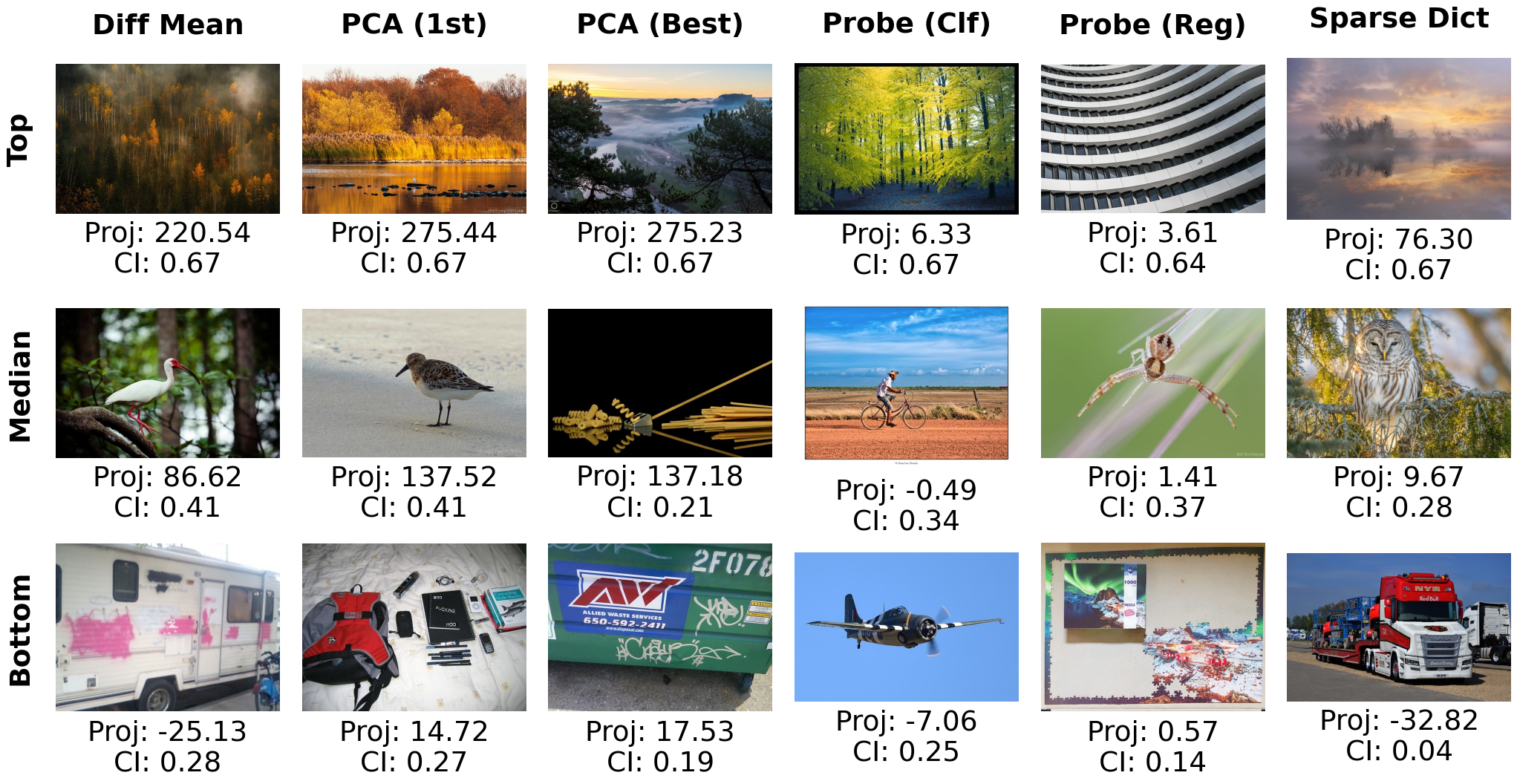}
  \end{center}\vspace{-0.5cm}
 \caption{\textbf{Representative images across terciles of concept vector activation.}
Each column corresponds to a distinct method for deriving concept vectors (Tab.~\ref{tab:concept_methods}). Final-layer hidden states were projected onto the learned concept vector for each method. Rows show images from the top, middle, and bottom terciles of projection scores, along with the projection value (Proj) and \gls{ci} score. High-scoring images typically depict natural landscapes with coherent composition, while low-scoring images tend to contain mundane or cluttered scenes.
}

  \label{fig:example_images}
\end{figure*}

\section{Methods}

\subsection{Qwen3-VL-8B} 

We employed Qwen3-VL-8B-Instruct, a transformer-based vision-language model that integrates text, image, and video understanding (\cite{bai2023qwen, bai2025qwen3vltechnicalreport}), using the pretrained version from Hugging Face (\cite{wolf2020transformers}). The model consists of a 27-layer Vision Transformer encoder (1152 hidden dimensions), a vision projector mapping visual features to the language model's embedding space, and a 36-layer Qwen3 language decoder (37 including embeddings as layer 0; 4096 hidden dimensions, 8.6B parameters total).
A simplified version of the architecture is illustrated in Fig.~\ref{fig:embedding_analysis} b.
We selected this model for its leading open-source performance and computational efficiency.

\subsection{Dataset and Common Interestingness Score}

\begin{figure*}[h!]
    \centering
    \includegraphics[width=\textwidth]{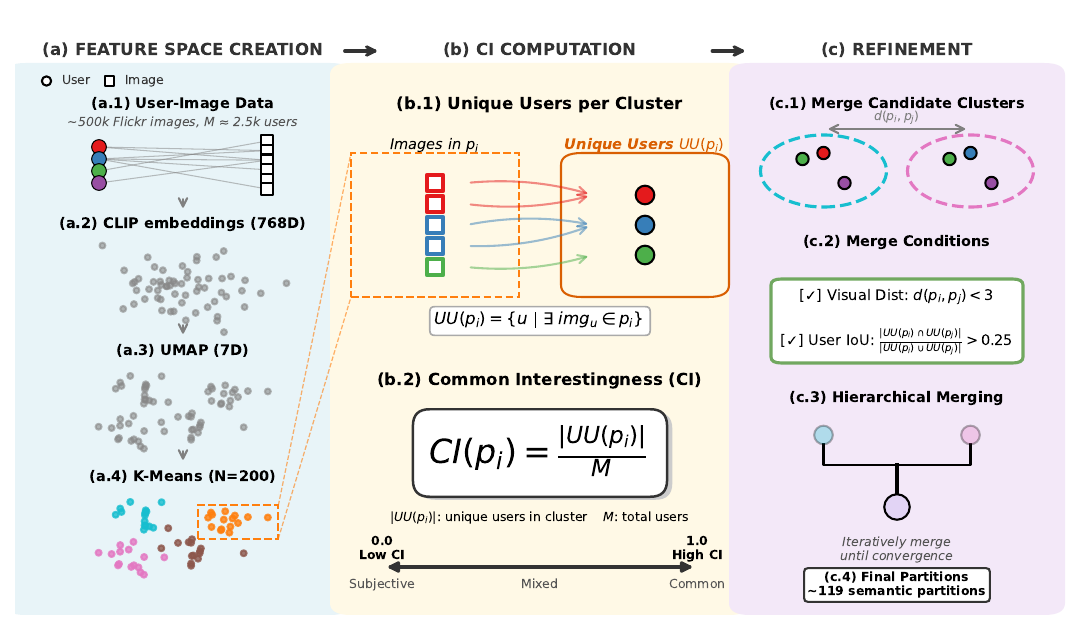}
\caption{\color{black}\textbf{CI Score Pipeline.}
We illustrate the Common Interestingness (CI) scoring method proposed by \citet{abdullahu2024commonly}. To move beyond the binary interesting/uninteresting distinction, they define CI as a data-driven continuum grounded in cross-user agreement.
(a)~CLIP embeddings (768D) of ${\sim}500$k Flickr images are reduced via UMAP and partitioned into $N{=}200$ initial clusters via $k$-means, providing semantic granularity fine enough to distinguish, e.g., 
cats from dogs which are subsequently merged, since ${\sim}60\%$ of users who like one also like the other.
(b)~For each partition $p_i$, $\text{CI}(p_i)$ measures the fraction of $M{\approx}2.5$k users who liked at least one image in that cluster: 
a score near $1.0$ reflects near-universal appeal (e.g., aesthetic landscapes), while a low score signals niche or personal interest. 
(c)~Partitions are iteratively merged via hierarchical clustering under the dual criteria of visual proximity and user overlap converging to 119 semantically coherent final partitions. Crucially, low-CI partitions are not uninteresting; they are subjectively interesting, appealing strongly to smaller communities.}
    \label{fig:ci_procedure}
\end{figure*}

{\color{black}
We used the Flickr dataset introduced by \citet{abdullahu2024commonly}, which contains images annotated with implicit user interaction data and associated \gls{ci} scores. The \gls{ci} score quantifies how broadly an image appeals across a diverse user population through the following process. First, all images are embedded using the \gls{clip} vision-language model, producing a rich semantic representation of each image's content. These high-dimensional embeddings are then reduced in dimensionality via \gls{umap} to improve clustering stability while preserving semantic structure. The resulting embedding space is partitioned into semantically coherent groups using K-Means clustering, such that images depicting similar content (for example, landscapes or portraits) fall into the same partition. User interactions are then mapped onto these partitions: for each partition, the number of unique users who liked at least one image within it is counted. Crucially, only one interaction per user per partition is considered, preventing prolific users from inflating the score. The \gls{ci} score of a partition is then defined as the fraction of all users who engaged with it. To reduce redundancy, partitions with similar \gls{clip} embeddings and comparable \gls{ci} scores are recursively merged via hierarchical clustering, yielding 119 final partitions. High \gls{ci} scores therefore indicate content with broad, cross-user appeal such as aesthetic landscapes, whereas low scores reflect niche or subjectively interesting content such as imagery from specific sports or hobbyist communities. Example images and corresponding \gls{ci} scores are shown in Fig.~\ref{fig:example_images} and the full procedure is illustrate in Fig.~\ref{fig:ci_procedure}.
}

{\color{black}
We note two structural limitations regarding the \gls{ci} metric derived from the Flickr dataset.
First, there is a shared representational geometry: \gls{ci} utilizes \gls{clip} embeddings to define semantic partitions, while Qwen3-VL's vision encoder is initialized from SigLIP-2. This architectural overlap may act as an inductive bias, easing the learning of \gls{ci}-predictive structure in the \gls{vit} and, indirectly, in the \gls{llm}, which processes the resulting visual tokens. Importantly, however, this does not threaten the validity of the \gls{ci} score as a behavioral ground truth: the score cannot be derived from \gls{clip} embeddings alone, as it is ultimately determined by real user interaction patterns.
Second, we acknowledge a label smoothing effect. \gls{clip} clustering aggregates distinct images into broad semantic partitions, assigning them a single averaged \gls{ci} score. This aggregation inherently dilutes fine-grained behavioral signals, smoothing over subjective visual nuances. Consequently, the \gls{ci} score serves as a coarse-grained metric of generalized appeal rather than a high-resolution measure of individual image preference.
}

{\color{black}
For all experiments, 4,000 images were subsampled using stratified random sampling across five bins spanning the \gls{ci} distribution, with the resulting indices saved and held fixed across all subsequent analyses. This subsample was partitioned into train, validation, and test sets (70/15/15) using a single stratified split preserving the \gls{ci} distribution.
Duplicate images were ruled out by computing pairwise cosine similarities between all embeddings, confirming that no identical pairs exist in the dataset.

Hidden states were then extracted at each transformer layer output for both \gls{vit} and \gls{llm} using the prompt ``Describe.''. The held-out test set (600 images) was used for all evaluations, including regression, correlation analyses, and representational similarity analysis.
}



\subsection{Computational Resources}
All experiments were conducted on a local cluster. \gls{ml} tasks were executed on NVIDIA A100 GPUs (40 GB VRAM), while non-\gls{ml} computations utilized CPU nodes equipped with AMD EPYC 7502 processors. Local debugging was performed on an NVIDIA RTX 4070 GPU using a 4-bit quantized 2B parameter model variant.

\subsection{Dimensionality Reduction and Generalized Discrimination Value}

Dimensionality reduction techniques were applied to project high-dimensional hidden representations into two-dimensional spaces for visualization and analysis. \gls{mds} (\cite{jaworska2009review, steyvers2002multidimensional}) was used to examine clustering structure in the vision transformer and language model components of Qwen, with representations color-coded by \gls{ci} score (\cite{abdullahu2024commonly}) using continuous gradients as well as three discrete binning strategies (binary median split, five sampling bins, and three trend groups). Complementary projections using \gls{pca}, \gls{tsne}, and \gls{umap} were performed to assess robustness (\cite{mcinnes2018umap, allaoui2020considerably, maaten2008visualizing}).

To quantify alignment between internal representations and \gls{ci}-based labels, the \gls{gdv} (\cite{schilling2021quantifying}) was computed for all projection methods and binning strategies. \gls{gdv} measures the normalized difference between intra- and inter-cluster distances (Eq.~\ref{GDVEq}), ranging from -1 (perfect clustering) to 0 (random distribution). Comparing \gls{gdv} values across methods allowed identification of projections that best preserved \gls{ci}-related structure in the representations. 

\begin{align}
\mbox{GDV}=\frac{1}{\sqrt{D}}\left[\frac{1}{L}\sum_{l=1}^L{\bar{d}(C_l)}\;-\;\frac{2}{L(L\!-\!1)}\sum_{l=1}^{L-1}\sum_{m=l+1}^{L}\bar{d}(C_l,C_m)\right]
 \label{GDVEq}
\end{align}
($D$: dimensionality of hidden representations, $L$: number of unique labels, $\bar{d}(C_l)$: mean intra-cluster distances, $\bar{d}(C_l,C_m)$: mean inter-cluster distances).
{\color{black}
To establish a statistical baseline, we conducted a permutation test with 1,000 trials. In each trial, \gls{ci} score labels were randomly shuffled across all images while keeping group sizes fixed, ensuring the null distribution reflected the same class proportions as the observed data. The \gls{gdv} was recalculated for each permutation in both the original and projected spaces, and empirical significance was determined by comparing the observed \gls{gdv} against this null distribution.
}

\subsection{Regression Head}

The regression head comprised a single linear layer applied to the final layer embeddings to predict \gls{ci} scores (see Fig.~\ref{fig:embedding_analysis} b). The model was trained using Adam with decoupled weight decay optimization with a cosine annealing learning rate schedule, incorporating a 10\% warmup period to stabilize early training. Training proceeded for up to 30 epochs, with early stopping employed (patience = 5 epochs) to prevent overfitting. To ensure robustness to outliers in the \gls{ci} score distribution, we used Huber loss with $\delta = 0.1$. Additional regularization was provided through gradient clipping and weight decay during optimization.

\begin{figure*}[h!]
    \centering
    \includegraphics[width=0.99\textwidth]{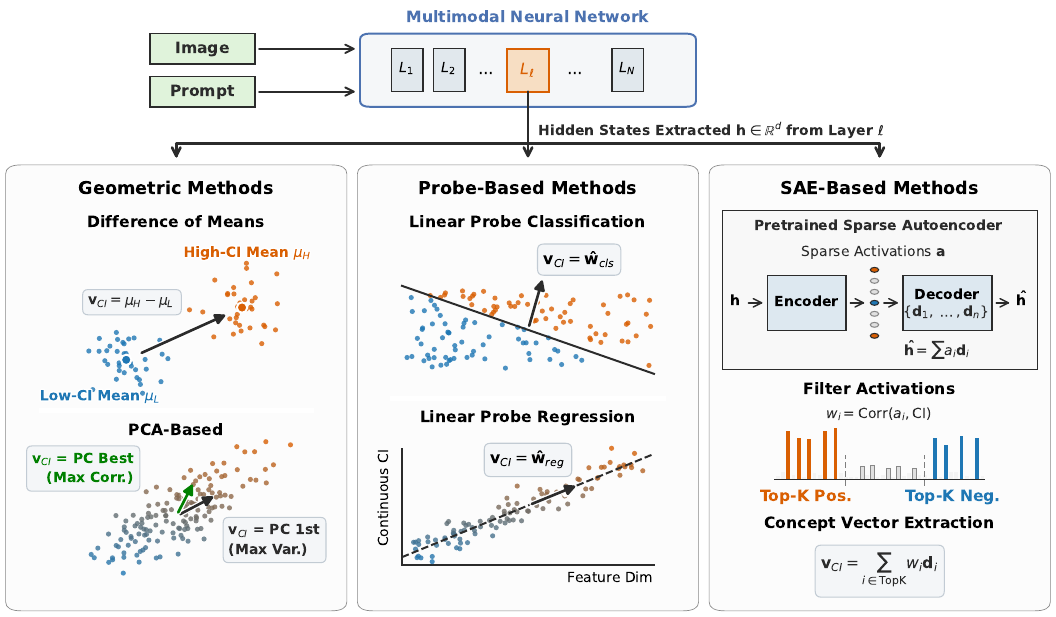}

    \caption{\color{black}\textbf{Concept vector extraction methodologies.} Hidden states $h \in \mathbb{R}^d$ are extracted from layer $\ell$ of a multimodal neural network. Six distinct approaches across three categories isolate the concept vector $v_{CI}$. Left: Geometric methods analyze representations directly. These include Difference of Means (computed between the top and bottom 20\% \gls{ci} samples) and \gls{pca}-based selection, which utilizes either the dominant variance direction (\gls{pca} 1st) or the direction most correlated with the concept (\gls{pca} Best). Middle: Probe-based methods learn predictive directions. Classification uses the decision boundary normal of a logistic regression model, while Regression defines the vector using weights learned via ridge regression. Right: \gls{sae}-based methods decompose dense states into sparse dictionary atoms. The final concept vector is a correlation-weighted sum of the 16 most positively and 16 most negatively correlated atoms.}
    \label{fig:concept_vector_extraction}

    \label{fig:concept_vectors}
\end{figure*}

\subsection{Concept Vectors}

Neural networks learn features corresponding to interpretable concepts (\cite{bereskamechanistic, rai2024practical}). Recent work suggests that such features are often well approximated by linear directions in activation space and may exist in superposition, with individual neurons contributing to multiple features (\cite{elhage2022toy}).

To test whether \gls{ci} manifests as a linear direction and to identify robust extraction methods, we evaluated six approaches for deriving candidate directions from layer-wise hidden states. We use the term concept vector to denote directions encoding \gls{ci}, generalizing the notion of \glspl{cav} (\cite{kim2018interpretability}). 
The methods fall into three categories: geometric, probe-based, and \gls{sae}-based approaches.

Geometric methods extract features directly from the underlying structure of the representations, functioning either independently or guided by \gls{ci} labels. Difference of Means (\cite{venhoff2025understanding}) computes the vector between mean embeddings of high- and low-\gls{ci} samples (top and bottom $20\%$). \gls{pca}-based approaches (\cite{marks2023geometry}) include \gls{pca} (1st), using the leading principal component, and \gls{pca} (Best), selecting the component (from the top 1000) most correlated with \gls{ci}, accounting for cases where \gls{ci} does not align with dominant variance.

Probe-based methods learn directions predictive of \gls{ci}. Linear Probe Classification trains a logistic regression classifier on median-split high/low-\gls{ci} samples, using the decision boundary normal as the concept vector (\cite{kim2018interpretability}). Linear Probe Regression applies ridge regression to predict continuous \gls{ci} scores, with learned weights defining the concept direction (\cite{frising2025linear}).

Finally, Sparse Dictionary Learning applies a \gls{sae} with 49,152 overcomplete features and top-K activation (K=16) to decompose representations into interpretable atoms (\cite{huben2023sparse}). The \gls{ci} concept vector is constructed from the 16 most positively and 16 most negatively correlated atoms as a correlation-weighted sum. {\color{black} All six methods are illustrated in Fig.~\ref{fig:concept_vectors}, } and Fig.~\ref{fig:example_images} shows example images with their projection values.

{\color{black}
To clarify their underlying mechanisms, our extraction methods can be classified along two classical dichotomies: model-based versus model-free, and supervised versus unsupervised. Geometric methods are strictly model-free: \gls{pca} (1st) is completely unsupervised, Difference of Means is supervised, and \gls{pca} (Best) combines unsupervised extraction with supervised selection. In contrast, both probe-based and dictionary approaches are model-based, requiring the optimization of auxiliary weights. The linear probes (both classification and regression) are explicitly supervised, optimizing directly to predict \gls{ci} labels. Finally, \glspl{sae} employ a sequential hybrid approach, pairing an unsupervised autoencoder decomposition with a supervised, post-hoc selection of \gls{ci}-correlated features. This is illustrated in Tab.~\ref{tab:concept_methods}

\begin{table}[h]
\centering
\renewcommand{\arraystretch}{1.5}
\begin{tabularx}{\columnwidth}{>{\bfseries}p{1.4cm} >{\raggedright\arraybackslash}X >{\raggedright\arraybackslash}X}
\toprule
 & \textbf{Unsupervised} & \textbf{Supervised} \\ 
\midrule
Model-\newline free & PCA (1st) & Difference of Means \\ 
 & \multicolumn{2}{c}{\begin{tabular}{@{}c@{}}\textit{PCA (Best): Unsup. Extr.} \\[-0.5ex] \textit{$\rightarrow$ Sup. Selection}\end{tabular}} \\ 
\midrule
Model-\newline based & \textit{--} & Linear Probes (Class. \& Reg.) \\ 
 & \multicolumn{2}{c}{\begin{tabular}{@{}c@{}}\textit{SAE: Unsup. Decomp.} \\[-0.5ex] \textit{$\rightarrow$ Sup. Selection}\end{tabular}} \\ 
\bottomrule
\end{tabularx}
\caption{\color{black}\textbf{Categorisation of Extraction Methods.}}
\label{tab:concept_methods}
\end{table}

}

\subsection{Pairwise Correlations and Representational Similarity Analysis}
As a first step, we analyzed the pairwise correlations between concept vector activations and the \gls{ci} score to examine their basic relationship. Subsequently, \gls{rsa} was used to systematically compare the structure of internal representations across the different components of the model and analytical techniques.
Following \citet{kriegeskorte2008representational, kriegeskorte2012representational}, we constructed \glspl{rdm} by computing pairwise dissimilarities between image representations in both the embedding and the concept vector activation space using Euclidean distances. We then calculated Pearson correlation coefficients between these \glspl{rdm} to assess the extent to which different model components and analytical methods preserve similar representational structures.
{\color{black}
Because the concept directions are derived using varying optimization strategies, \gls{rsa} provides a robust framework to compare the structural similarity of the representations they induce.
}

\begin{figure*}[h!]
  \begin{center}
    \includegraphics[width=\textwidth]{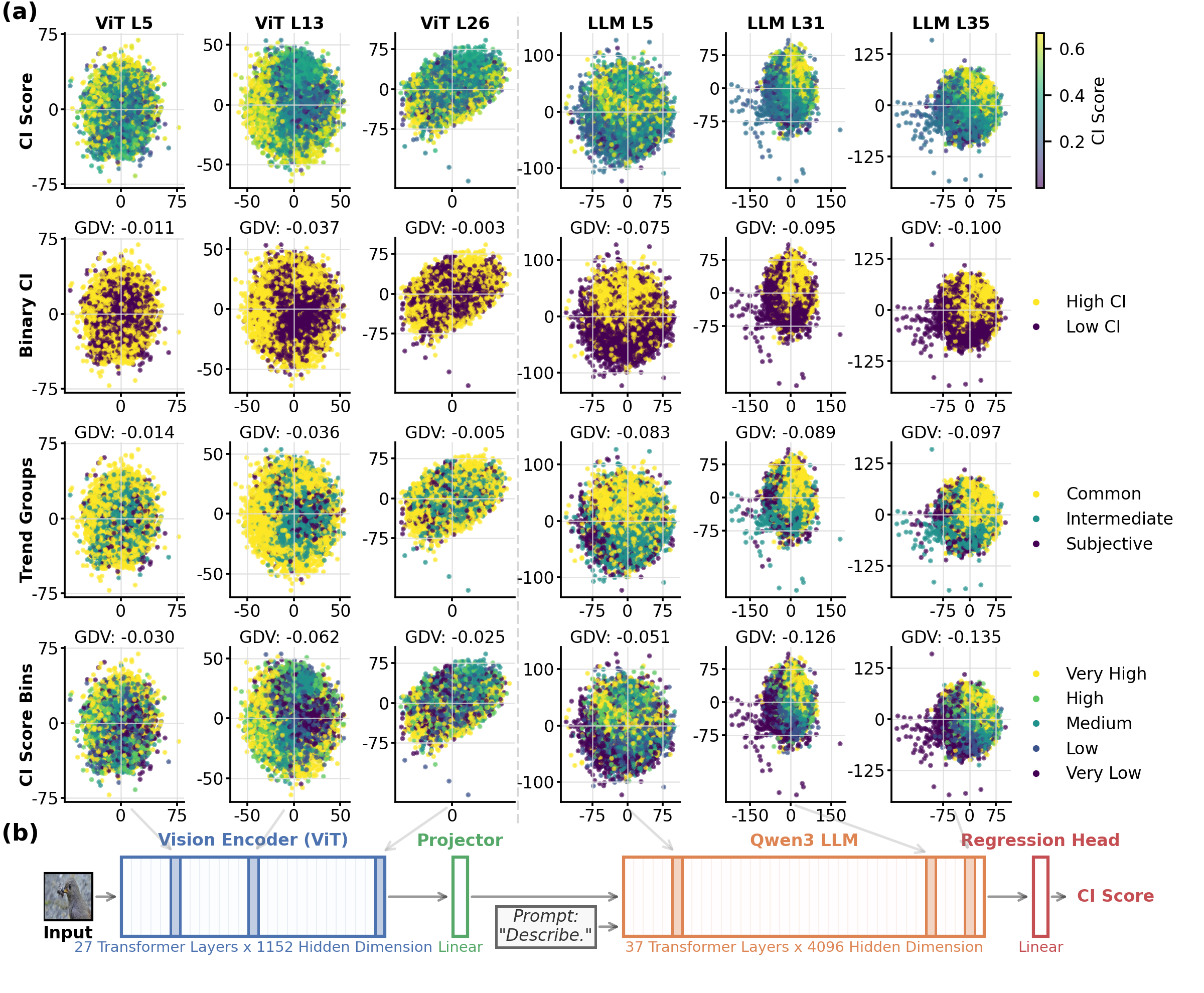}
  \end{center}\vspace{-0.5cm}
\caption{
\textbf{Multidimensional scaling projections of embedding spaces across Vision Transformer (ViT) and Language Model (LLM) layers including a simplified Qwen3-8B VL architecture} (a) 2D MDS projections of embeddings from three ViT layers and three \gls{llm} layers colored-coded by: continuous \gls{ci} scores (row 1), binary \gls{ci} groups via median split (row 2), trend groups from \citet{abdullahu2024commonly} (row 3), and quintile \gls{ci} score bins (row 4). \gls{gdv} quantify clustering quality, with lower values indicating better separation between high and low \gls{ci} images. (b) Architecture diagram highlighting the analyzed layers within the vision encoder and Qwen3 \gls{llm} components of the vision-language model.
}
\label{fig:embedding_analysis}
\end{figure*}

\section{Results}

\subsection{Visual interestingness representations}

Visual interestingness, quantified here through the \gls{ci}, is explicitly represented in the model’s highest-level embeddings. \gls{ci} scores can be reliably predicted from final-layer representations using a simple linear readout, indicating that information related to human visual interest forms a structured and linearly accessible component of the learned representation space and that multimodal models implicitly capture population-level regularities of human attention and preference without an explicit objective targeting interestingness.

To quantify this effect, a linear regression head was trained on final-layer embeddings to predict \gls{ci} scores. The model showed strong predictive performance on a held-out test set (Tab.~\ref{tab:metrics}), explaining $57.5\%$ of the variance (R$^2$) with low prediction error (RMSE = 0.112, MAE = 0.085). Pearson and Spearman correlations confirmed robust linear and monotonic relationships between predicted and observed \gls{ci} scores ($r$ = 0.776, $\rho$ = 0.764; both $p < 10^{-10}$); given the negligible difference between measures, subsequent analyses report Pearson correlations for consistency.

\begin{table}[h]
\centering
\small
\begin{tabular}{lc}
\hline
\textbf{Metric} & \textbf{Value} \\
\hline
RMSE & 0.112 \\
MAE & 0.085 \\
R² & 0.575 \\
Pearson $r$ & 0.776 \\
Spearman $\rho$ & 0.764 \\
\hline
\end{tabular}
\caption{\textbf{Regression head performance metrics on the test set.} The model demonstrates strong predictive accuracy for \gls{ci} scores, with high correlations and low prediction error across multiple evaluation metrics.}
\label{tab:metrics}
\end{table}

\subsection{Interestingness emerges across model hierarchy}

Having established that visual interestingness is encoded in the model’s final-layer representations, we next examined where along the model hierarchy this information emerges. We find that \gls{ci}-related structure is not uniformly present across layers, but develops progressively during processing. While early visual layers show only weak separation between images with high and low \gls{ci}, this structure becomes increasingly pronounced at intermediate stages and reaches maximal discriminability in deeper language model layers. This pattern indicates that visual interestingness emerges through successive stages of abstraction rather than being determined solely by low-level visual features.

To examine how \gls{ci} is represented across the model hierarchy, \gls{mds} was applied to embeddings from selected \gls{vit} layers (L5, L13, L26) and language model layers (L5, L31, L35) (Fig.~\ref{fig:embedding_analysis}a).
{\color{black}All 4,000 images were used for the MDS projections, as they require no training.}
Clear \gls{ci} gradients emerged in intermediate vision layers and became increasingly pronounced across language model layers, where images with high and low \gls{ci} scores were progressively separated. In contrast, this structure was weaker in early and late vision layers. These observations were confirmed by \gls{gdv} analyses, which showed weak clustering in early visual processing (\gls{vit} L5: \gls{gdv} = -0.011 to -0.030), stronger separation at intermediate vision layers (\gls{vit} L13: \gls{gdv} = -0.037 to -0.062), and reduced discriminability at later vision stages (\gls{vit} L26: \gls{gdv} = -0.003 to -0.025). After the transition to the language model, \gls{ci}-related structure strengthened monotonically across layers (\gls{llm} L5: \gls{gdv} = -0.051 to -0.083; \gls{llm} L31: \gls{gdv} = -0.089 to -0.126; \gls{llm} L35: \gls{gdv} = -0.097 to -0.135), indicating that representations of visual interestingness are progressively refined along the hierarchy and reach maximal discriminability in deeper language model layers.

\begin{figure*}[h!]
  \begin{center}
    \includegraphics[width=0.9\textwidth]{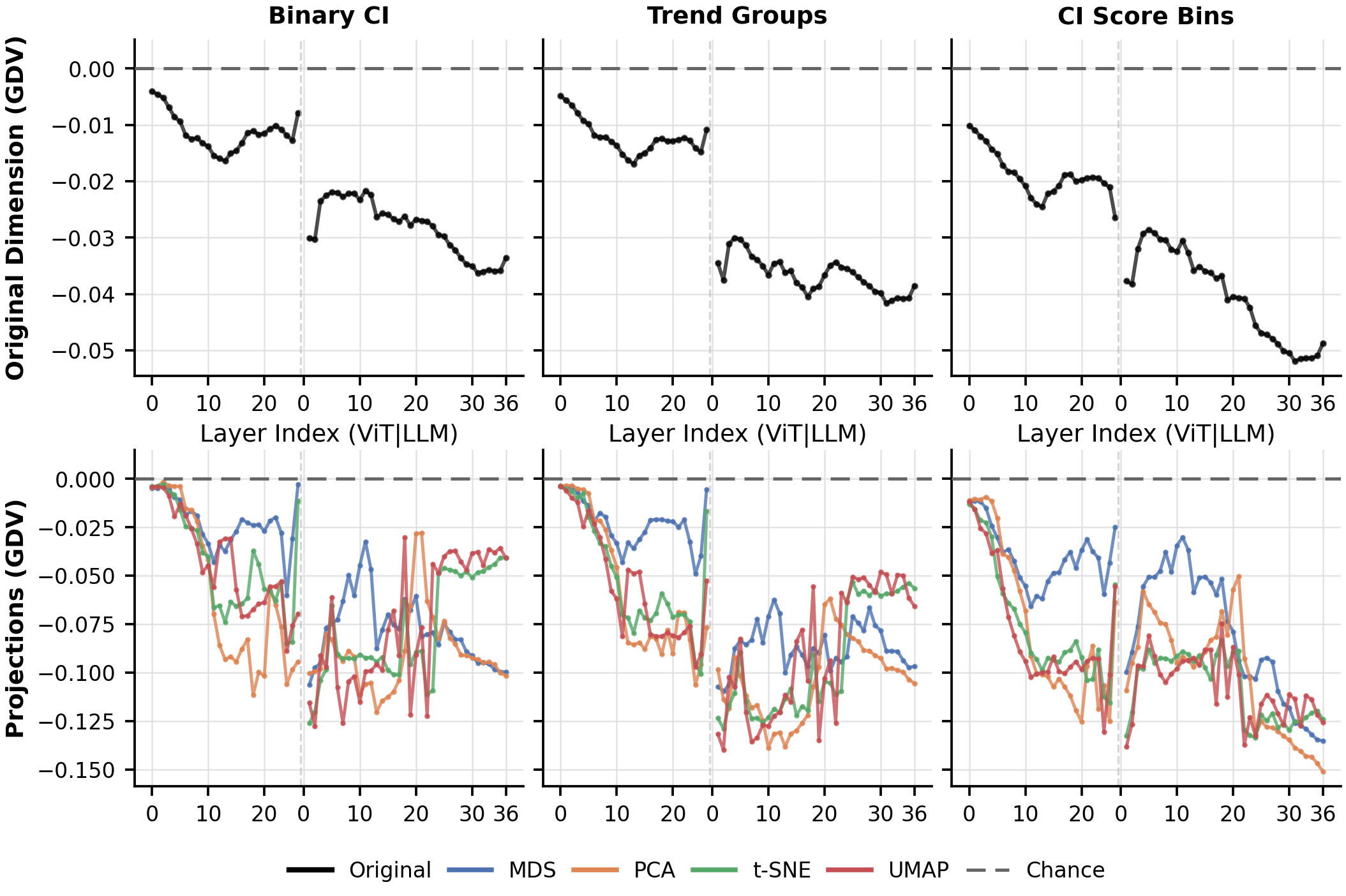}
  \end{center} \vspace{-0.5cm}

\caption{
\textbf{Layerwise Generalized Discrimination Values (GDV) across the full model hierarchy.} 
Top row: \gls{gdv} computed in the original high-dimensional embedding space for each layer using three grouping strategies (Binary \gls{ci}: median split; Trend Groups: commonly, intermediate and subjectively interesting; \gls{ci} Score Bins: quintiles). Lower values indicate better separation between \gls{ci} groups. Bottom row: Layerwise \gls{gdv} computed after dimensionality reduction using \gls{mds}, \gls{pca}, \gls{tsne}, and \gls{umap} projections. Vertical dashed lines demarcate the transition from \gls{vit} (layers 0-26) to \gls{llm} (layers 0-36). Both the high-dimensional embeddings and all projection methods generally capture the progressive refinement of \gls{ci} representations from vision to language layers, though projections show varying fluctuations. {\color{black} Chance level is at 0 indicated by the horizontal dashed line.}
}
\label{fig:gdv_analysis}
\end{figure*}

Building on the layer-specific patterns above, \gls{ci}-related structure was systematically quantified across the full model architecture by computing layerwise \gls{gdv} for all vision transformer and language model layers using three grouping strategies (Fig.~\ref{fig:gdv_analysis}). 
{\color{black} All 4,000 images were used for \gls{gdv} calculation, as it requires no training.}
In the original high-dimensional space, \gls{gdv} values showed a U-shaped pattern within the vision encoder, with weak clustering in early layers, stronger separation at intermediate vision layers (\gls{vit} L13), and reduced discriminability toward the final vision layer. After the transition to the language model, clustering improved substantially, reaching a global minimum around \gls{llm} L31. This pattern was consistent across grouping strategies, with \gls{ci} score bins yielding the strongest clustering. Dimensionality reduction methods (\gls{mds}, \gls{pca}, \gls{tsne}, \gls{umap}) produced stronger absolute separation but preserved the same overall trend of progressively increasing \gls{ci} discriminability from vision to language processing, despite method-specific variability across layers.
{\color{black}
Permutation testing confirmed that all observed \gls{gdv} values were significantly below chance across every layer, grouping strategy, and projection method (all $p < 0.001$, 1,000 iterations) achieving a different order of magnitude compared to the random label. This rules out the possibility that the observed \gls{ci}-related clustering arose from random label assignment.
}

\begin{figure*}[h!]
  \begin{center}
    \includegraphics[width=0.97\textwidth]{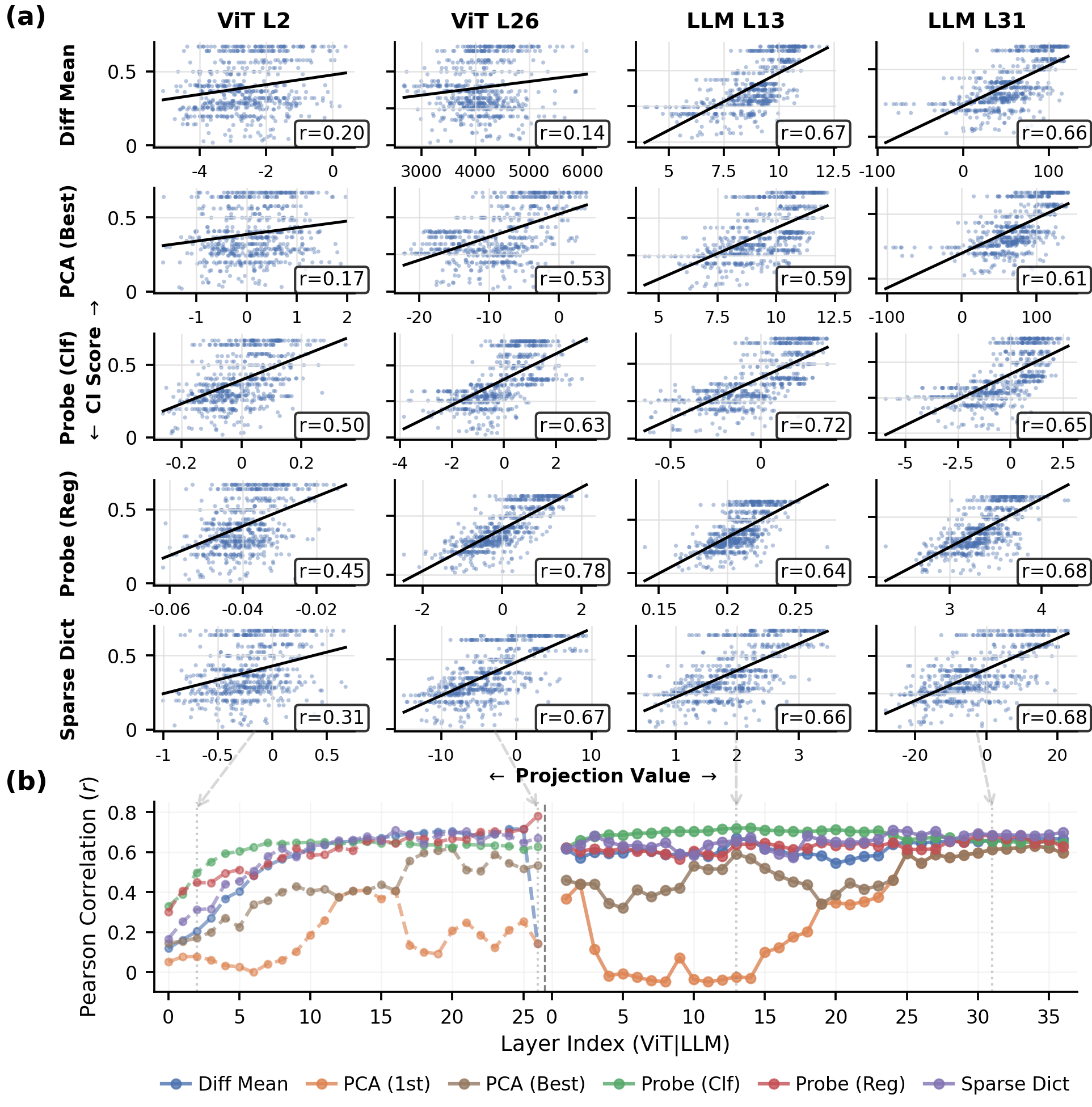}
  \end{center}\vspace{-0.5cm}
 \caption{ 
\textbf{Concept vector projections correlate with Common Interestingness (CI) scores across model layers.} (a) Scatter plots showing the relationship between concept vector projection values and \gls{ci} scores for representative layers from early and late vision processing (ViT L2, L26) and middle and late language processing (\gls{llm} L13, L31) across five extraction methods: Difference of Means, PCA (Best), Probe (Clf), Probe (Reg), and Sparse Dictionary Learning. Each point represents a single image, with Pearson correlation coefficients (r) displayed. Black lines indicate linear fits. (b) Pearson correlations between concept vector projections and \gls{ci} scores across all model layers for six methods, including \gls{pca} (1st) which extracts the primary direction of variance. Vertical dashed line separates ViT (layers 0-26) from \gls{llm} (layers 0-36). Correlations increase during early processing (layers 0-10) before stabilizing at high values (r > 0.6) for most methods, with PCA (1st) showing markedly lower performance through intermediate layers before converging with other methods in the deepest \gls{llm} layers.}
\label{fig:concept_vectors}
\end{figure*}

\subsection{Interestingness forms a linear concept direction}

To characterize how visual interestingness is represented internally, we examined whether \gls{ci} corresponds to a consistent direction in representation space. Across analyses, interestingness emerged as a largely linear concept direction that became increasingly stable across model depth, with different extraction approaches identifying strongly aligned directions in deeper layers. To quantify this effect, interpretable concept directions were extracted across all model layers using six complementary methods, including geometric approaches, learned probes, and sparse dictionary learning (Fig. \ref{fig:concept_vectors}a,b).
{\color{black}
Concept vectors were fit on the training set (2,800 images) and evaluated on the held-out test set (600 images). }
All methods identified \gls{ci}-aligned directions, with correlations increasing during early processing and stabilizing in later layers. Probe-based methods and sparse dictionary learning showed the most consistent performance, while geometric methods exhibited greater variability, particularly in late vision layers. \gls{pca}-based analyses further indicated that \gls{ci} is not generally aligned with the dominant variance axis. Despite methodological differences at intermediate stages, performance converged in deeper \gls{llm} layers (L31–36: r = 0.6–0.7), indicating that \gls{ci} emerges as a stable and readily accessible organizational feature of the representation space.

\begin{figure*}[h!]
  \begin{center}
    \includegraphics[width=\textwidth]{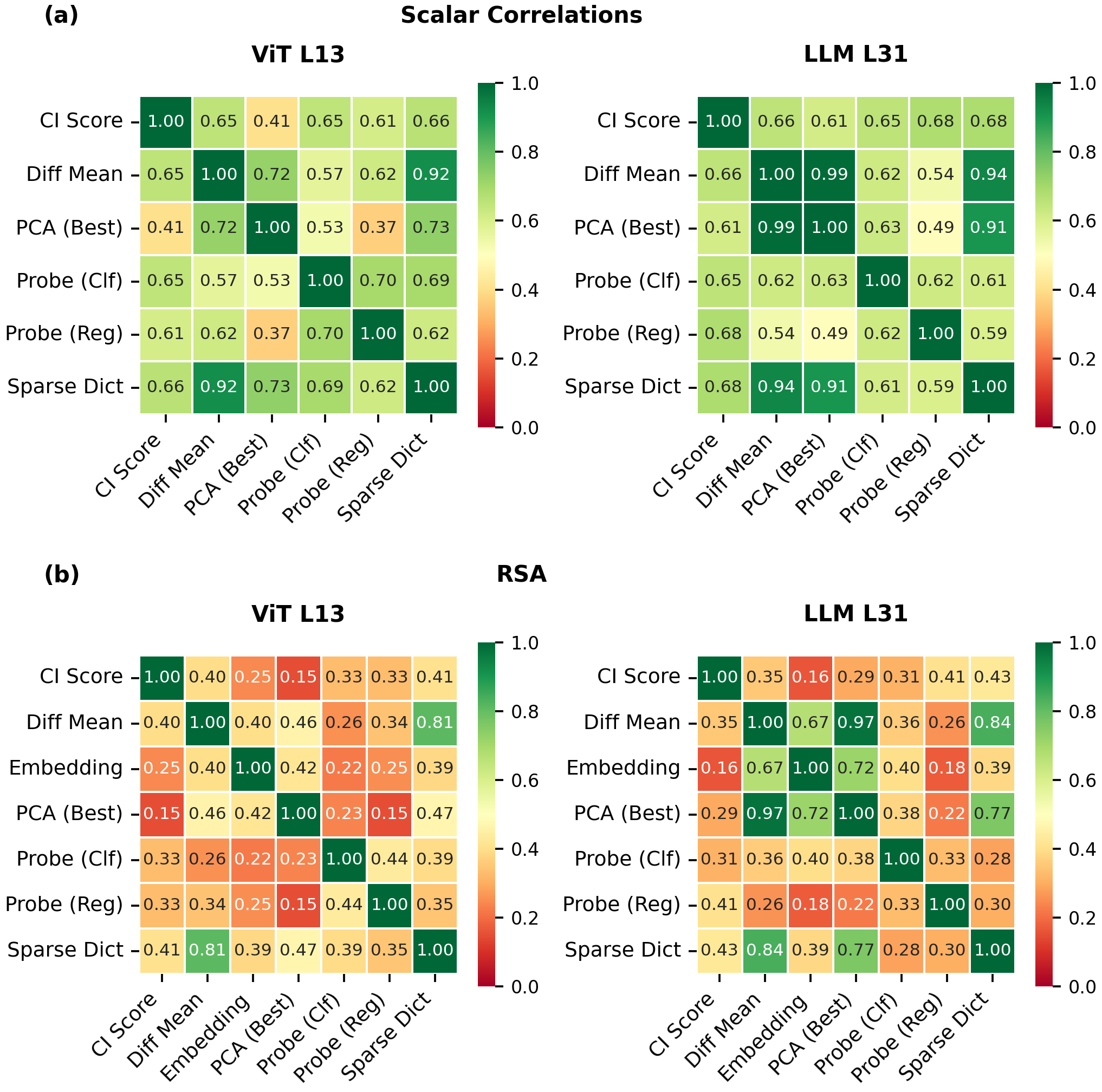}
  \end{center}\vspace{-0.5cm}
  \caption{
   \textbf{Pearson correlations and Representational Similarity Analysis reveal agreement across concept vector methods.} Pairwise comparisons between concept vector extraction methods and \gls{ci} scores at intermediate vision (ViT L13) and late language processing (\gls{llm} L31). (a) Pearson correlations between projection values onto each concept vector and \gls{ci} scores. (b) Representational Similarity Analysis (RSA) computed as Pearson correlations between \glspl{rdm}, where each \gls{rdm} is constructed from Euclidean distances between samples in the space of concept vector projections. The "Embedding" row represents distances in the original high-dimensional embedding space. Cell values indicate correlation coefficients, with darker green indicating stronger positive correlation. While different methods achieve similar correlations with \gls{ci} (a), their pairwise distance structures show more variability (b).
  }
  \label{fig:corr_rsa}
\end{figure*}

\subsection{Convergence across analytical methods}

Having established that visual interestingness forms a consistent concept direction, we next examined whether this structure depends on the analytical method used for its extraction. Across methods, we observed substantial convergence, with different approaches recovering highly similar \gls{ci}-related structure, particularly in deeper language model layers, indicating that interestingness reflects a robust property of the model’s internal geometry rather than a method-specific artifact.

To quantify this agreement, concept vectors derived from different extraction approaches were compared using pairwise Pearson correlations and \gls{rsa} at representative layers (\gls{vit} L13, \gls{llm} L31; Fig.~\ref{fig:corr_rsa}). 
{\color{black}
Both were performed on the held-out test set (600 images). }

Correlations showed strong agreement across methods, increasing in deeper language model layers (r~=~0.37–0.92 at \gls{vit} L13; r~=~0.49–0.99 at \gls{llm} L31), with unsupervised methods exhibiting particularly high mutual correlations (r~=~0.91–0.99).

\subsection{Emergence of interest in semantic representations}

Finally, we examined how \gls{ci}-related information is organized within the model’s broader representational geometry. While different analytical approaches converged on similar \gls{ci}-discriminative directions, they captured partially distinct aspects of the underlying structure, indicating that visual interestingness emerges as a distributed property embedded within higher-level semantic representations rather than as an isolated feature.

Consistent with this view, \gls{rsa} of representational distance structures revealed greater divergence across methods, with correlations between \glspl{rdm} ranging from r~=~0.15–0.81 at \gls{vit} L13 and r~=~0.16–0.97 at \gls{llm} L31. The original high-dimensional embedding space showed moderate alignment with concept vector distance structures, strongest for unsupervised methods, indicating that different extraction approaches recover similar \gls{ci}-discriminative directions while capturing complementary aspects of its high-dimensional organization.

\section{Discussion}

This study demonstrates that, despite the absence of any explicit objective to model human interest, human-derived \gls{ci} scores are systematically encoded in the internal representations of a multimodal transformer. Visual interestingness is not only reflected in model outputs, but emerges as a structured and linearly accessible component of the learned representation space. Together, the results show that signals derived from large-scale human behaviour become embedded in multimodal representations rather than arising solely at the level of downstream decisions.

The present work should be distinguished from prior research on aesthetic prediction \citep{talebi2018nima}, CLIP-based probing \citep{radford2021learning}, and preference learning \citep{ouyang2022training}. Aesthetic prediction models and related approaches typically aim to directly predict subjective quality, appeal, or engagement from images, often optimizing supervised objectives on curated datasets. Similarly, CLIP-based probing studies focus on extracting or predicting semantic or perceptual attributes from pretrained representations, while preference learning frameworks explicitly model human choices or rankings through dedicated training procedures. In contrast, our study does not aim to improve predictive performance or learn a model of visual preference. Instead, we take a post hoc, representation-centric perspective and analyze whether and how a population-level measure of visual interestingness is already reflected in the internal structure of a large multimodal model that was not explicitly trained for this objective. Our findings therefore complement existing work by shifting the focus from prediction to representation, suggesting that signals related to shared human interest can emerge as accessible and structured properties of pretrained multimodal embeddings, rather than requiring explicit supervision or task-specific optimization. While part of this structure may be facilitated by shared representational priors between pretrained vision encoders and the CI construction pipeline, the consistent emergence across model components suggests that it is not reducible to a trivial embedding overlap. Crucially, this perspective allows us to dissociate behavioral alignment from representational structure, addressing the question of whether human-relevant signals are merely recoverable at the output level or are systematically reflected in the internal geometry of the model.

A central finding is that the representation of visual interestingness develops progressively across the model hierarchy. Alignment with \gls{ci} is comparatively weak in early vision transformer layers and increases substantially after the transition into the language model, indicating that visual interestingness is not primarily determined by low-level perceptual properties but emerges through successive stages of abstraction. The observation that interestingness forms a consistent and largely linear concept direction further suggests that this structure reflects an organizational feature of the representation space rather than a localized or method-specific effect. At the same time, representational similarity analyses indicate that \gls{ci}-related information remains distributed, with different analytical perspectives capturing complementary aspects of how interest is embedded within the model’s high-dimensional geometry.

{\color{black}

Further clarifying how this interestingness is organized within the model's semantic space, the distinct performance trajectory of the \gls{pca} (1st) method demonstrates that \gls{ci} does mainly align with the dominant axis of variance in the later \gls{llm} layers. The delayed convergence of \gls{pca} (1st) compared to targeted methods like linear probes and sparse dictionaries indicates that the model allocates its primary representational capacity to other computational demands, such as structural understanding or instruction following. Human interest operates in these layer as a distributed, orthogonal property embedded within this broader geometry, requiring targeted extraction techniques to be reliably isolated.

The analysis reveals a critical distinction between identifying interestingness as an isolated feature versus using it as an organizational principle for the representational space. As demonstrated by the concept vector projections, the linear direction corresponding to CI is successfully isolated as early as the intermediate vision layers.
This early convergence is consistent with the structural overlap between the ViT's SigLIP-2 initialization and the CLIP-derived CI partitions, which provides a favorable geometric substrate for extracting CI-associated visual features. Nevertheless, linear separability is not guaranteed by initialization alone: the CI score reflects real human behavioral responses, a signal that is absent from the encoder's initialization and must therefore be recoverable from the model's learned representations of visual content.
Additionally, it is particularly notable that the predictive performance of these concept vectors in the LLM layers matches, and often exceeds, the peak performance observed in the ViT. Despite lacking the structural bias that inherently favors the vision encoder, the language model proves equally or more capable of linearly isolating the concept of visual interestingness.

Furthermore, while our concept vectors successfully isolate the concept in both components, the \gls{gdv} analysis reveals a fundamental structural transformation across the model hierarchy. Despite early linear separability, the overall structural clustering of images by CI remains weaker within the vision encoder and only peaks in the deeper LLM layers. This discrepancy indicates that in the ViT, visual interestingness is merely one of many detectable perceptual features which are easily found by targeted probes but not dominant enough to cluster the overall geometry. Upon entering the LLM, the representational space is reorganized. The language backbone leverages these extracted visual features to structurally cluster the images, elevating visual interestingness from a localized perceptual direction into a defining semantic category.

While predictive performance reaches high values relatively early in the model hierarchy and remains stable across the \gls{llm} layers, it encounters a persistent ceiling around r=0.7 to 0.8. We interpret this generalized plateau not as a limitation of the model's representational capacity, but as a resolution ceiling imposed by the ground truth metric {\color{black} itself. The \gls{ci} score is fundamentally a partition-level measure: it assigns a single score to a broad semantic category of images rather than to individual images. Because the score is derived by aggregating user interactions across all images within a CLIP-based cluster, fine-grained visual differences between images in the same partition are entirely lost. Two images of landscapes, one a mediocre snapshot and another a professionally composed photograph, receive an identical \gls{ci} score simply by virtue of belonging to the same semantic cluster. This coarse granularity means that even a model capable of extracting highly nuanced visual representations cannot be fully validated against such a target, as the ground truth itself lacks the resolution to reward or penalize fine-grained distinctions.}

{\color{black} A natural and necessary next step is therefore to move towards per-image interestingness measures, grounded in direct human judgments rather than aggregated interaction data. Beyond granularity,} semantic interest is highly fluid and dependent on observer context{\color{black}, meaning that even a per-image score remains a simplification}. A promising {\color{black} complementary} direction for future research is the use of contextual prompting. By introducing specific user personas or situational contexts into the {\color{black} model's} prompt prior to extracting representations, researchers could simulate a distribution of subjective evaluations, moving beyond a single {\color{black} static} score to dynamically model how context shifts the geometry of visual interest.

}

In multimodal systems trained on image–text data derived from real-world human usage, visual representations become embedded within grounded semantic spaces linking perception to meaning and context (semantic grounding; see e.g., \citet{lyre2024understanding, pulvermuller2018case, koelbl2025prediction}). The stronger alignment with common interest observed in language model layers therefore suggests that visual interestingness emerges primarily at the level of semantic representation rather than perceptual processing alone.

More broadly, the convergence of human behavioural patterns and internal representations in multimodal transformers hints at shared organizing principles underlying attention and interest. Although human brains and transformer models differ fundamentally in architecture and learning dynamics, both appear to shape internal representations according to relevance, informativeness, and behavioural significance. This parallel suggests that learning systems may converge toward common regions of representational space — potentially reflecting platonic representations (\citet{huh2024platonic, metzner2025core, metzner2025platonic, sole2026cognition}). Identifying these hidden regularities may ultimately provide a unifying framework for understanding how attention and interest emerge across natural and artificial cognitive systems.


\section{Acknowledgments}
This work was funded by the Deutsche Forschungsgemeinschaft (DFG, German Research Foundation): KR\,5148/5-1 (project number 542747151), KR\,5148/10-1 (project number 563909707) and GRK\,2839 (project number 468527017) to PK, and SCHI\,1482/6-1 (project number 563909707) to AS. Furthermore, the research was supported by the EELISA European University program to PK and HG.

\printbibliography

\end{document}